# Role of Artificial Intelligence in Detection of Hateful Speech for Hinglish Data on Social Media

Ananya Srivastava [1,3][0000-0002-3495-1854], Mohammed Hasan[1,3][0000-0002-1270-1758], Bhargav Yagnik[1,3][0000-0002-0705-703X], Rahee Walambe [2][0000-0003-1745-5231] * and Ketan Kotecha [2][0000-0003-2653-3780]

[1] Symbiosis Institute of Technology, Pune, India
[2] Symbiosis Centre for Applied Artificial Intelligence
[3] These authors contributed equally to this work
*Corresponding Author:rahee.walambe@scaai.siu.edu.in

**Abstract.** Social networking platforms provide a conduit to disseminate our ideas, views and thoughts and proliferate information. This has led to the amalgamation of English with natively spoken languages. Prevalence of Hindi-English code-mixed data (Hinglish) is on the rise with most of the urban population all over the world. Hate speech detection algorithms deployed by most social networking platforms are unable to filter out offensive and abusive content posted in these code-mixed languages. Thus, the worldwide hate speech detection rate of around 44% drops even more considering the content in Indian colloquial languages and slangs. In this paper, we propose a methodology for efficient detection of unstructured code-mix Hinglish language. Fine-tuning based approaches for Hindi-English code-mixed language are employed by utilizing contextual based embeddings such as ELMo (Embeddings for Language Models), FLAIR, and transformer-based BERT (Bidirectional Encoder Representations from Transformers). Our proposed approach is compared against the pre-existing methods and results are compared for various datasets. Our model outperforms the other methods and frameworks.

## 1 Introduction

In the upfront of our social lives, lies the huge platform of social media sites. Social media penetration in India is growing very rapidly with currently over 29 percent of India's population using social media [5]. Due to the rise in usage of social media, contrasting ideology and hateful material on the Internet has escalated. An individual's liberty to free speech is prone to exploitation and hate can be conveyed in terms of specific or particular speech acts or disjunctive sets of speech acts –for example, advocating hatred, insulting or defaming, terrorizing, provoking discrimination or violence, accusation [12]. Hate speech has an adverse effect on the mental health of individuals [11]. Twitter have seen about 900% increase in hate speech during COVID-19 pandemic [15]. YouTube reported about 500 Million hate comments from June 2019 to September 2019 [4]. Hate speech has been recognized as a growing concern in the society and numerous automated systems have been developed to identify and avert it.

With the recent progress in pre-trained models across different language models and support for Hindi and other languages, better results could be achieved in Hinglish hate speech classification.

### 1.1 Previous Work

One of the earliest works in hate speech recognition is reported in [22] which extracted rule-based features to train their decision tree text classifier. Number of studies have been carried out particularly within the context of social media [6,19,8]. In [27], authors surveyed various application of Deep Learning algorithms to learn contextual word embeddings for classification of tweets as sexist or racist or none using deep learning. Mozafari [16] reported better performance of Convolutional Neural Network (CNN) with BERT on [8] dataset. Hateful text classification in online textual content has also been performed in languages like Arabic [17] and Vietnamese [10].

Various approaches and methods for detection of abusive/hateful speech in Hinglish have been reported in the literature. Hinglish is a portmanteau of the words in Hindi and English combining both in a single sentence.[28] Now with increase in the number of youths using this mixed language, not only in urban or semi-urban areas, but also to the rural and remote areas through the social media, Hinglish is achieving the status of a vernacular language or a dialect. Sinha [21], attempted to translate Hinglish to standard Hindi and English forms. However, due to shallow grammatical analysis, the challenge of polysemous words could not be resolved. [14] created an annotated dataset of Hindi English Offensive Tweets (HEOT) split into three labels: Hate Speech, Non-Offensive Speech and Abusive Speech. Ternary Trans CNN model was pretrained on tweets in English [8] followed by retraining on Hinglish tweets. An accuracy of 0.83 was reported on using this transfer learning approach. Multi-Input Multi-Channel Transfer Learning using multiple embeddings like Glove, Word2vec and FastText with CNN-LSTM parallel channel architecture was proposed by [13] that outperformed the baselines and naïve transfer learning models. If the text classification approaches designed for English language are applied to such a code-mixed language like Hinglish, it fails to achieve the expected accuracy. Some of these standard models were applied on the Hinglish dataset and the results obtained have been reported in section 5.

In English Language there was an extensive work on Hate speech datasets but there were no datasets focusing on Indian Social media texts, so we have focused on developing a hate speech classifier based on Indian audience. The previous methods explained by researchers in Hinglish language did not implement pre-trained transformers embeddings considering the barrier of Hindi pre-trained language models. With the advances in pre-trained language models like Multilingual BERT [23] and support for Hindi and other languages we were able to implement Models like BERT, ELMo and FLAIR to exceed the previously obtained results in English and Hinglish. In summary, the contributions of this work are:

- Design of custom web-scraping tools for different social media platforms and development of an annotated Hate speech dataset specific to Indian audiences for both English and Hinglish languages.
- Development of separate generic pre-processing pipelines for Hinglish and English languages which are validated on benchmark datasets.
- Demonstration of use of BERT for Hinglish Embeddings using pre-trained multi-lingual weights.
- Demonstration and validation of a novel approach based on the ELMO and implementation of FLAIR framework for Hinglish Text classification.

The paper is divided into following sections. In section 2, the data collection and annotation scheme for both English and Hinglish datasets has been illustrated. In the section 3, classification systems including BERT, ELMO, FLAIR have been summarized and the pipelines followed for both English and Hinglish datasets have been discussed in section 4. In section 5, results are presented and compared. In the last Section, the conclusion and future scope is discussed.

## 2     Corpus Creation and Annotation

The datasets provided by Davidson [8] and in Hinglish by Bohra [3] and [13] have been used for validation and demonstration of the proposed approach. In addition to these datasets, primary part of this work was focused on collection and creation of the datasets us for both languages with the intention of encompassing a wider spectrum of issues across various verticals of the society.

The data was collected from Twitter, Instagram, and YouTube comments dated from November 2019 till February 2020 associated with domains such as political, cultural, gender, etc. The major portion of the corpus was from trending news headlines and hashtags which went viral on social media platforms as they may gather more hate.

Custom web-scraping tools were designed for YouTube and Instagram using selenium [20] while "Twitter_scrapper" was used to scrape data from Twitter [24].The comments that were collected were saved into CSV file format.

### 2.1     Annotations

The comments collected were non-homogeneous i.e. they were a mix of all three languages namely, English, Hindi and Hinglish. Hence the segregation of the comments into their respective languages was important as different languages required different pre-processing pipelines. This was carried out by assigning separate labels for each language. The comments were then annotated based on the definition of hate speech stated by the United Nations [26]. The labels of the two classes were "Hate" and "Not-Hate". Table 1 shows the examples of hate speech and subsequent annotations from our dataset.

**Table 1.** Examples of Hate Speech and subsequent annotations

|  | English | Hinglish |
|---|---|---|
| Generic | "The most illiterate, ill-manered, psychopath Just barks whatever shit out of his mouth @euler12 😂" | "Bikao kutta sala.. 2 kodi ka insaan.. Tujhe sataye hue ke baddua lagenge harami 😡" |
| Political | We have a stupid government, divide and rule is their motto .#BJP | "Tuje toh kutta bhi nahi bol sakta wo bhi wafadar hota hai tu toh desh gad-dar hai sale madarchod" |
| Gender Based | "Hoes what do you expect. Make money by showin body" | "Chutiya orat...kisse aajadi chahiye tuje ...randapa krne ki aajadi bol ?" |
| Religion Based | "As Muslim will never leave their ideology of killing non-Muslims and now other communities have as well given up on living together peacefully." | "madarchod musalman rape hindu girls" |

Annotations were performed by the authors who are proficient Hindi as well as English speakers. The inter annotator agreement [7] was observed to be 0.89. There were around 8000 (32.5%-YouTube, 47.9%-Instagram, 19.5%-Twitter) comments collected which were then divided into 2 datasets, 1683 comments in Hinglish text, while 6160 comments in English. The comments in Hindi were transliterated to Hinglish using a Unidecode [25]. Followed by dataset creation, the development of classifiers has been discussed in the next section.

## 3 Methodology

Textual data in its raw form cannot be understood by computers and hence embeddings are used in NLP to convert text into multi-dimensional vectors. The recent developments in state-of-the-art language models, have proven to be extensive in giving excellent results with small amount of training data. In this section we demonstrate the of use BERT for English and Hinglish Embeddings using pre-trained weights and validate a novel approach based on the ELMO and implementation of FLAIR framework for Hinglish Text classification.

### 3.1 BERT (Bidirectional Encoder Representations from Transformers)

BERT is a deep bi-directional transformer-based language representation model [9] which is designed finetuning with shallow neural networks to obtain state of the art models. $BERT_{base-uncased}$ ($BERT_{BU}$) and $BERT_{multilingual-uncased}$ ($BERT_{MU}$) were used for this work. $BERT_{BU}$ contains 768 hidden layers and is trained on English Wikipedia and the Book Corpus. BERT$_{MU}$ is similar but trained on lower-cased text in top 102 Languages in Wikipedia. So, this becomes useful when encoding text in Hinglish Language. BERT was then finetuned using our datasets.

### 3.2 ELMo (Embeddings from Language Models)

ELMo [18] embeddings developed by AllenNLP include both word level contextual semantics and word level characteristics. It is character based, i.e., the model forms a vocabulary of words that are not present in the vocabulary and captures their inner structure. It uses bi-directional LSTMs to create word representations according to the context of the words they are used in. The forward pass contains words before the target token while the backward pass contains words after the target token. This forms the intermediate word vector which acts as input to the next layer of Bidirectional language model. The final output (ELMo) is the weighted sum of the intermediate word vectors and the input word vector.

### 3.3 FLAIR

Flair [2] library consists of models such as GloVe, ELMo, BERT, Character Embeddings, etc. This interface allows stacking-up of different embeddings which gives a significant increase in results. "Flair Embeddings" [1] which are unique to the Flair library uses contextual string embeddings that capture latent syntactic-semantic information and are contextualized based on neighboring text leading to different embeddings for the same words depending on the context. Flair includes support for pre-trained multilingual embeddings including 'hi-forward' and 'hi-backward' for Hindi.

The standard CNN [29] and Bi-LSTM frameworks are used with the above specified embeddings. In the next section, we implement distinctive preprocessing approaches along with the above-mentioned methodologies.

## 4 SYSTEM DESIGN

### 4.1 Data Pre-processing

Separate generic pre-processing pipelines for Hinglish and English languages were constructed and validated on the benchmark datasets.

**English**
The social media data tends to be casual and informal, leading to a decrease in the ability of a language model to understand the corpus, hence performing extensive preprocessing on the data became necessary. Fig. 1 represents the diagrammatic pipeline for preprocessing in English language. The initial step was expansion of contractions and abbreviations into their standard notation. Common pronouns, conjunctions, articles and prepositions in English vocabulary generally add no contextual meaning to a sentence and are ignored by search engines thus were removed from the corpus. Removal of URLs, hashtags, mentions and punctuations was also carried out as a part of pre-processing. The emojis were replaced with the text denoting its meaning.

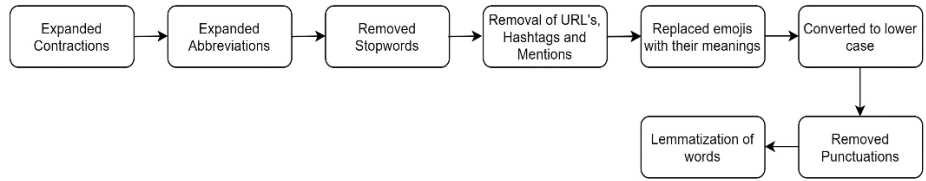

**Fig. 1.** Pre-processing Pipeline for English Dataset

Lastly, to map the words into their root form, WordNetLemmatizer (NLTK) was applied. Table 2 shows an example of the preprocessing done on the English corpus.

**Table 2.** Example of Pre-processing technique for English dataset

| Text | Pre-processed text |
| --- | --- |
| @amitshah You can't change the minds of such small minded people who are stuck in the past, they just don't understand logics.#IndiaAgainstCAA 👿 | you cannot change mind small minded people stuck past understand logic face-withsymbolsonmouth |

**Hinglish**

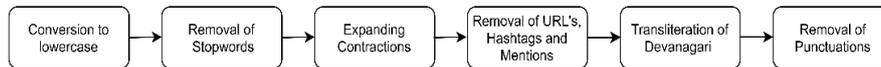

**Fig. 2.** Pre-processing Pipeline for Hinglish Dataset

Prior to performing the pre-processing techniques to Hinglish data, the comments were manually cleaned to remove ambiguity, perform spell check, removal of Hinglish specific stopwords and repetitive comments. For this the English stopwords list was appended and commonly used Hinglish words along with variations in their spellings were added to the list. For example, words like *teko, terko, tujhe,* etc. were added to the Hinglish stopwords list.

**Table 3.** Example of Pre-processing technique for Hinglish dataset

| Text | Pre-processed text |
| --- | --- |
| @narendramodi मेरा देश BHAI hate ni pyar phailata ha or jo pyar se nhi manta wo use ache se samjhate hain!!😂 https://twitter.com/4948747235330 | meraa desh hate ni pyar phailata pyar nhi manta ache samjhate hain |

The text was then converted into lowercase as shown in Figure 2. The Hinglish Data contained a few words in Devanagari script that had to be converted to Roman Script. For this purpose, Unidecode library was used to transliterate the Hindi text to Hinglish. Table 3 shows an example of the preprocessing done on the Hinglish corpus.

### 4.2 Fine-Tuning Approaches

**BERT$_{BU}$ + CNN for English Dataset**
The processed text was then converted to tokens using WordPiece Tokenizer which has a vocabulary of 30523 unique keywords. The outputs were then truncated into 100 tokens along with padding for smaller sequences. The tokens were then used to generate weights of dimensions 1x768 from BERT$_{BU}$. These weights were then fine-tuned using the CNN architecture proposed by [29] that consisted of 6 filters, 2 filters of sizes 2,3 and 4 each followed by a Max Pooling Layer and a Softmax Layer. The learning rate was set to 1e-3 and the model was trained on NVIDIA DGX Station with 32 GB RAM.

**BERT$_{MU}$ + CNN for Hinglish Dataset**
The processed text in Hinglish was converted into tokens using the WordPiece Tokenizer but this time using a different vocabulary file for BERT$_{MU}$ which has a vocabulary of more than 1M keywords from 102 Languages. The outputs were then truncated to 75 tokens. Sequences lesser than it were padded and then the weights were generated by using the BERT$_{MU}$ language model which were fine-tuned to specific task of hate speech detection using above mentioned CNN architecture with a learning rate of 1e-4 for the HOT dataset [13] (3500 samples) while learning rate of 1e-3 was used for the dataset annotated by the authors, combined with HOT [13] and Hate speech dataset [3] (9000 samples).

**ELMo and MLP for Hinglish Dataset**
For implementing ELMo, ELMo model 2 was imported using Tensorflow Hub. The processed text was taken as raw input vectors for ELMo embeddings and intermediate vectors were computed. The concatenation of outputs of both the layers was carried out to have one vector for each word with a size of 1024. The learning rate was set to 1e-5 with Adam Optimizer. The training process was carried out on Google Colaboratory GPU.

**Flair**
The pre-processed data was split into train (80%), test (10%) and dev (10%) sets following the convention of flair text classifier with text and labels represented in the form:                label__<class_n> <text>
Further two approaches were used under flair:

*Flair-stacked (WordEmbeddings$_{hi}$ + FlairEmbeddings$_{hi+forward}$ + FlairEmbeddings$_{hi+backward}$)*

The input split above is embedded over the stacked pre-trained multilingual embeddings of "WordEmbeddings$_{hi}$", "FlairEmbeddings$_{hi+forward}$" and "FlairEmbeddings$_{hi+backward}$" which were specifically chosen for Hinglish based applications. The weights were then fine-tuned to explicit detection using BiLSTM with a learning rate of 1e-5 and Adam optimizer. It was trained on Google Colab GPU Tesla P100.

*Flair (BERT + BiLSTM)*

Two different embeddings were used in this approach. One was BERT$_{BU}$ while the other was BERT$_{MU}$. The input split mentioned above is embedded over two separate pre-trained multilingual embeddings of "BERT$_{BU}$" and "BERT$_{MU}$". The weights from both the embeddings were then fine-tuned to explicit detection using an BiLSTM with a learning rate of 1e-5 and Adam optimizer and trained on Google Colab GPU.

The above-mentioned approaches were used to train the models on the collected datasets. The results of the same have been discussed in the following section.

## 5      RESULTS AND DISCUSSION

We evaluate the outcomes of different fine-tuning approaches on their datasets and compare them with other baseline datasets. Table 4 summarizes the different results obtained from various fine-tuning approaches for English mentioned in section 4. In English, two datasets were considered for this purpose: Davidson[8] dataset which was trained on BERT$_{BU}$ + CNN approach resulting in an accuracy of 94% while the English Hate Dataset collected by the us was trained on ELMO and BERT$_{BU}$+CNN both yielding a similar accuracy of 73%. The considerable variation between the datasets is mainly because of the shift in domain as the text in Davidson dataset is mainly generic while the dataset collected by us is event-driven i.e. it contains comments based on political, gender-specific, religious events which in-turn disseminates different hate based contextual meaning that may not be directed by any specific abusive words.

**Table 4.** Results for English Datasets

| Dataset | Model | Accuracy | Recall | F1 Score |
|---|---|---|---|---|
| Davidson [8] Dataset | BERT$_{BU}$ + CNN | 94% | 0.94 | 0.93 |
| English Hate dataset | ELMO | 73% | 0.73 | 0.71 |
| | BERT$_{BU}$ + CNN | 73% | 0.73 | 0.70 |

The pipelines used in English, were applied on the HOT dataset in Hinglish and the results obtained are summarized in Table 5 and show that the English pipelines were ineffective and led to acquiring lower accuracy on Hinglish dataset given the complex nature of Hinglish corpus. Thus, designing a different pipeline for Hinglish dataset

9was necessary. Table 6 depicts that after applying a different pipeline specific to Hinglish dataset, significantly better results were achieved.

**Table 5.** Results of English pipeline applied to Hinglish Dataset

| Dataset | Model | Accuracy | Recall | F1 Score |
|---|---|---|---|---|
| HOT dataset [13] (Hinglish) | $BERT_{BU}$ + CNN | 83% | 0.83 | 0.83 |
| | ELMO | 80% | 0.80 | 0.79 |

**Table 6.** Results for Hinglish Datasets based on Hinglish Pipeline

| Dataset | Model | Accuracy | Recall | F1 Score |
|---|---|---|---|---|
| HOT dataset [13] (Hinglish) | FLAIR Embeddings $_{hi}$ + Flair $_{hi\text{-}forward+hi\text{-}backward}$ | 81% | 0.84 | 0.91 |
| | FLAIR $BERT_{MU}$ + bi-LSTM | 84% | 0.86 | 0.92 |
| | ELMO + MLP | 85% | 0.85 | 0.85 |
| | $BERT_{MU}$ + CNN | 86% | 0.86 | 0.86 |
| | FLAIR $BERT_{BU}$ + Bi-LSTM | 88% | 0.89 | 0.94 |
| Hinglish Hate dataset + HOT [13] | ELMO + MLP | 71% | 0.71 | 0.70 |
| | $BERT_{MU}$ + CNN | 82% | 0.82 | 0.82 |
| | FLAIR $BERT_{MU}$ + Bi-LSTM | 82% | 0.84 | 0.91 |
| Hinglish hate dataset + HOT [13] + [3] dataset | ELMO + MLP | 67% | 0.67 | 0.66 |
| | BERT | 67% | 0.67 | 0.67 |
| | FLAIR $BERT_{MU}$ + Bi-LSTM | 73% | 0.79 | 0.87 |

The results obtained using $BERT_{MU}$ were at par with the results of $BERT_{BU}$ in terms of accuracy for Hinglish datasets. But $BERT_{MU}$ can be considered preferable as it has a broader vocabulary including words from many languages and hence it was able to tokenize words in a more contextual form. For example, "kaam karna he" would be generated as "[CLS] Ka ##am ka ##rna He [SEP]" using $BERT_{BU}$ while using $BERT_{MU}$ it would be "[CLS] kaam karna he [SEP]". Also, this leads to loss of context in case of $BERT_{BU}$ as it contains words from the English vocabulary and tokenizes out-of-vocabulary words into multiple sub-word tokens.

Table 7 and 8 compare the obtained results with the baseline results observed previously on Davidson dataset [8] in English and HOT dataset [13] in Hinglish.

**Table 7.** Comparison for accuracies in Davidson Dataset [8]

| Method | F1-Score |
|---|---|
| Waseem [27] | 0.89 |
| Mozafari [16] | 0.92 (LSTM) |
| | 0.92 (CNN) |
| $BERT_{BU}$ -CNN | 0.93 |

**Table 8.** Comparison for accuracies in HOT Dataset

| Method | F1- Score |
| --- | --- |
| Multi-Channel CNN-LSTM Architecture Mathur [13] | 0.89 |
| FLAIR BERT$_{MU}$ + Bi-LSTM dataset | 0.92 |
| ELMO | 0.85 |
| BERT$_{MU}$ + CNN | 0.86 |
| FLAIR BERT$_{BU}$ + Bi-LSTM | 0.94 |

F1-score of 0.93 was achieved on Davidson dataset [8] using the BERT$_{BU}$ fine-tuned with CNN which exceeds the previous results obtained from [16]. For this task around 30 K samples from the Davidson dataset [8] were used.

Along with the English dataset, the best results for Hinglish were yielded using BERT$_{MU}$ + CNN and FLAIR BERT$_{BU}$ + Bi-LSTM approach, attaining F1-scores of 0.86 and 0.94 respectively, which exceeds the baseline results obtained by [13]. Fig. 3 depicts the confusion matrices for BERT$_{BU}$ on Davidson [8] Dataset and ELMO applied on Hinglish dataset [21] respectively.

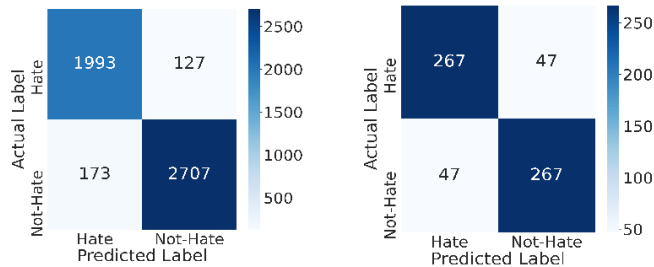

**Fig. 3.** Figure 3. (Left) Confusion matrix for BERT with Davidson dataset (Right) Confusion Matrix for ELMO applied on Hinglish dataset

## 6  CONCLUSION

In this work we have proposed to improve the hate speech detection in English as well as Hinglish languages with specific focus on the social media. Language modelling and text classification in English language is comparatively a well explored area, However, for a code-mixed language like Hinglish, which is very common and popular in India, the detection of hate speech is highly challenging. We collected the English and Hinglish datasets manually from various online avenues. Fine-tuning based approaches for Hindi-English code-mixed language are employed by utilizing contextual based embeddings such as ELMO, FLAIR and transformer-based BERT. We compared our proposed approach with various existing methods and demonstrated that our methods are promising and outperform the other approaches. Our major contribution lies in designing the annotated hate speech dataset in Hinglish, developing a pre-processing pipeline which is generic in nature and devolvement and validation of deep learning based approaches for detection of hate speech in Hinglish text.

### 6.1 Future Scope

Many application-driven pipelines can be designed for Hinglish embeddings as well as classifiers given the advancement in the field of NLP for multilingual domain and the scope of transfer learning for further accuracy. Enhanced approaches include changing the architecture of neural network classifier or using deeper/ multi-layer neural networks with larger corpus for Hindi-English code mix language. The project can also be extended to multiclass classification under degree of hate or as labelled in Davidson [8] dataset that includes classes like toxic, severe-toxic, obscene, threat, insult and identity-hate rather than bi-class i.e. Hate and Not Hate. The research presented in this paper can be extended to other code-mix local + English languages from any part of the world. We hope that the dataset and the experimental results encourage further research in multi-lingual domain as well as Hate-speech recognition.